\documentclass[journal]{IEEEtran} 
\usepackage{graphicx}
\usepackage{xurl}
\usepackage{hyperref}
\usepackage{textcomp} 
\usepackage{ulem} 
\usepackage{tabularx}
\usepackage[font=small,skip=0pt]{caption}
\usepackage[font=small,skip=0pt]{subcaption}
\usepackage{lscape}
\usepackage{tikz}
\usetikzlibrary{decorations.pathreplacing} 
\usetikzlibrary{shapes.geometric}
\usepackage{dblfloatfix} 
\usepackage{grffile} 
\usepackage{float} 
\usepackage{xspace} 
\usepackage{xcolor}
\usepackage{soul}
\usepackage{algorithm2e}
\usepackage{amsmath}

\hypersetup{
    colorlinks = true,
    linkcolor=red,
    citecolor=red,
    urlcolor=blue,
    linktocpage 
}

\newcommand{\banosdataset}[0]{Banos \textit{et al}\xspace}
\newcommand{\recofitdataset}[0]{Recofit\xspace}
\newcommand{\covtypedataset}[0]{Covtype\xspace}

\newcommand{\mondrianforest}[0]{Mondrian forest\xspace}

\newcommand{\mondriantrees}[0]{Mondrian trees\xspace}

\definecolor{revcolor}{RGB}{176,27,119}
\sethlcolor{revcolor}

\title{{Dynamic Ensemble Size Adjustment for
Memory Constrained Mondrian Forest}}

\author{Martin Khannouz and Tristan Glatard\\
Department of Computer Science and Software Engineering\\ Concordia University, Montreal, Quebec, Canada}

\begin{document}
\maketitle
\begin{abstract}
		Supervised learning algorithms generally
		assume the availability of enough memory  to
		store data models during the training and test
		phases. However, this assumption is
		unrealistic when data comes in the form of
		infinite data streams, or when learning
		algorithms are deployed on devices with
		reduced amounts of memory. Such memory
		constraints impact the model behavior and
		assumptions. In this paper, we show that under
		memory constraints, increasing the size of a tree-based
		ensemble classifier can worsen its performance. In particular,
		we experimentally show the existence of an
		optimal ensemble size for a memory-bounded
		\mondrianforest on data streams and we design
		an algorithm to guide the forest toward that
		optimal number by using an estimation of
		overfitting. We tested different variations
		for this algorithm on a variety of real and
		simulated datasets, and we conclude that our
		method can achieve up to 95\% of the
		performance of an optimally-sized
		\mondrianforest for stable datasets, and can
		even outperform it for datasets with concept drifts.
		All our methods are implemented in the
		OrpailleCC open-source library and are ready
		to be used on embedded systems and connected
		objects.
\end{abstract}
\begin{IEEEkeywords}
Mondrian Forest, Trimming, Concept Drift, Data
Stream, Memory Constraints.
\end{IEEEkeywords}

\section{Introduction}
\label{sec:introduction}
Supervised classification algorithms mostly assume the availability of abundant
memory to store data and models. This is an issue when processing data streams
--- which are infinite sequences by definition --- or when using memory-limited
devices as is commonly the case in the Internet of Things. We focus on the
\mondrianforest, a popular online classification method. Our ultimate goal is to
optimize it for data streams under memory constraints to make it compatible with
connected objects.

The \mondrianforest is a tree-based, ensemble, online learning method with
comparable performance to offline Random Forest~\cite{mondrian2014}. Previous
experiments highlighted the \mondrianforest sensitivity to the ensemble size in
a memory-constrained environment~\cite{khannouz2022}.
Indeed, introducing a memory limit to the ensemble also introduces a trade-off
between underfitting and overfitting. On the one hand, low tree numbers make room for
deeper trees and increase the risk of overfitting. On the other hand, large
tree numbers constrain tree depth due to the memory limitation and increase the
risk of underfitting. Depending on the memory limit and the data distribution, a given ensemble
size may either overfit or underfit the dataset. The goal of this paper is to address this
trade-off by adapting the ensemble size dynamically.

In summary, this paper makes the following contributions:
\begin{enumerate}
		\item Highlight the existence of an optimal tree count in memory-constrained
				\mondrianforest;
		\item Propose a dynamic method to optimize the tree count;
		\item Compare this method to the \mondrianforest with a
				optimally-sized tree count.
\end{enumerate}



\section{Materials and Methods}
All the methods presented in this section are implemented in the OrpailleCC
framework~\cite{OrpailleCC}. The scripts to reproduce our experiments are
available on GitHub at
\url{https://github.com/big-data-lab-team/benchmark-har-data-stream}.

In this section, we start by presenting background on Mondrian Forests
(\ref{sec:mondrian-forest} and \ref{sec:data-stream-mondrian-forest}), then
presents the main contribution of the paper, namely dynamically adjusting the
ensemble size of a memory-constrained \mondrianforest
(\ref{sec:dynamic-tree-count-optimization}, \ref{sec:comparison-test},
\ref{sec:mean-variance}, \ref{sec:adding-tree}), then describes the
experimental evaluation framework (\ref{sec:datasets},
\ref{sec:evaluation-metric}).

\subsection{Mondrian Forest}
\label{sec:mondrian-forest}
The \mondrianforest~\cite{mondrian2014} is an ensemble method that aggregates
Mondrian trees. Each tree recursively splits the feature space,
similar to a regular decision tree.  However, the feature used in the split and
the value of the split are picked randomly. The probability to select a feature
is proportional to its range, and the value for the split is
uniformly selected in the range of the feature.
In contrast with other decision trees,
the Mondrian tree does not split leaves to introduce new nodes. Instead, it
introduces a new parent and a sibling to the node where the split occurs. The
original node and its descendant are not modified and no data point is moved to
that new sibling besides the data points that initialized the split.
This approach allows the Mondrian tree to introduce new branches to internal
nodes. This training algorithm does not rely on labels to build the tree,
however, each node maintains counters for each label seen. Therefore, labels can
be delayed, but are needed before the prediction. In addition to the counters,
each node keeps track of the range of its feature which represents a box 
containing all data points. A data point can create a new branch only if it is
sorted to a node and falls outside of the node's box.

\subsection{Mondrian Forest for Data Stream Classification}
\label{sec:data-stream-mondrian-forest}
The implementation of \mondrianforest presented in~\cite{mondrian2014,
mondrian_implementation_1} is online because trees rely on potentially all the
previously seen data points to grow new branches. To support data streams, the
\mondrianforest has to access data points only once as the dataset is assumed to
be infinite in size.

The work in~\cite{khannouz2022} describes a Data Stream \mondrianforest with a
memory bound. The ensemble grows trees from a shared pool of nodes and the trees
are paused when there is no node left. This work also proposed out-memory strategies
to keep updating the statistics for the trees without creating new branches. In
particular, this work recommend using the Extend Node strategy when the memory is
full, a strategy where statistics of the node boxes are automatically extended
to fit all data points, and the counters automatically increased.

The attributes of a node are an array
for counting labels that fall inside a leaf, and two arrays 
$lower\_bound$ and $upper\_bound$ that define a box of the node. The Extend Node
strategy automatically increases the counter of the label in the leaf and
automatically adjusts $lower\_bound$ and $upper\_bound$ so the new data point
fits inside the box.

Having a shared pool of nodes for the ensemble has a direct impact on the number
of trees. As mentioned before, having more trees limits the tree depth and may lead to 
underfitting, whereas having less trees increases the risk of overfitting.

\subsection{Dynamic Tree Count Optimization}
\label{sec:dynamic-tree-count-optimization}
Algorithm~\ref{alg:training-forest} describes the function that trains the
forest with a new data point and dynamically adjusts the ensemble size.The main
idea is to compare pre- and post-quential errors to decide whether or not to
adjust the forest size.

\begin{algorithm}
\SetAlgoLined
\LinesNumbered
\SetKwProg{Fn}{Function}{ is}{end}
\KwData{f = a \mondrianforest}
\KwData{x = a data point}
\KwData{l = the label of data point x}
\Fn{train\_forest(f, x, l)}{

		predicted\_label = f.test(x);\\
		prequential.update(predicted\_label, l);\\
		\For{i}{
				train\_tree(i, x, l);\\
		}
		predicted\_label = f.test(x);\\
		postquential.update(predicted\_label, l);\\
		post\_metric = postquential.metric();\\
		pre\_metric = prequential.metric();\\
		\If{post\_metric $\tilde{>}$ pre\_metric}{
				trim\_trees(f);\\
				add\_tree(f);\\
		}
}
		\caption{Training function for a data stream \mondrianforest with a dynamic
		ensemble size.}
\label{alg:training-forest}
\end{algorithm}

The forest evaluates prequential statistics, meaning that it
predicts the new data point label before using it for training (line 2-3).
After training, the forest evaluates 
postquential statistics (lines 7-8). The prequential accuracy is widely used as accuracy
approximation on a test set for data
streams~\cite{issues_learning_from_stream}. Here we introduce the postquential
accuracy, where we test after training, to simulate the accuracy on a training
set.

To find the number of trees that maximize prediction performance, we initialize
the \mondrianforest with a single tree, therefore with a high risk of
overfitting. The idea is to test for overfitting by comparing prequential and
postquential accuracies and add a tree in case the forest is deemed to overfit. 

A model that overfits is defined as a model that performs significantly better
on the training set than on the test set. Therefore the problem of detecting
overfitting becomes a statistical testing problem between the training and test
accuracies.

While overfitting is commonly detected by comparing the performance of the
classifier on the training and test set, detecting underfitting for
memory-constrained data stream models remains very challenging. Consistently,
Algorithm~\ref{alg:training-forest} is written to only support tree addition.
Should an underfitting criterion be available in the future,
Algorithm~\ref{alg:training-forest} could easily be adapted to support tree
removal.

Overall, Algorithm~\ref{alg:training-forest} can be adjusted with three
components: the comparison test (line 11), the type of pre/postquential
statistics used (line 3 and 8), and the method to add trees (line 12-13).

\subsection{Comparison Test}
\label{sec:comparison-test}
In Algorithm~\ref{alg:training-forest}, the update process determines if it
needs to add a tree based on a comparison between the
prequential and the postquential accuracies. If both accuracies are
significantly different, the forest is deemed to overfit and thus, the algorithm
adds a tree to compensate.

There are different methods to test the statistical difference between two
accuracies and we experiment with four in this paper: the sum of
variances, the t-test, the z-test, and the sum of standard deviations.

Notations for the following equations include: $\mu_{pre}$ and $\mu_{post}$ the
mean of respectively the prequential and postquential accuracies, $\sigma_{pre}^2$ and
$\sigma_{post}^2$ the variance of respectively the prequential and postquential
accuracies, $n$ the size of the sample, $\mu$ and $\sigma^2$ respectively the
mean and variance of $\mu_{post} - \mu_{pre}$.

\subsubsection{Sum of Variances}
Equation~\ref{eqt:sum-var} shows the sum of variances as a comparison test. The
two accuracies are different when the distance it is higher than the square root of the
prequential and postquential variances.

\begin{equation}
		\label{eqt:sum-var}
		\mu_{post} - \mu_{pre} > \sqrt{\sigma_{post}^2 + \sigma_{pre}^2}
\end{equation}

\subsubsection{T-test}
Equation~\ref{eqt:t-test} describes the t-test used to compare the prequential
and postquential accuracies. We apply a one-sample t-test where we check if
$\mu$ is different from 0 with a 99\% confidence~\cite{t-test}.

\begin{equation}
		\label{eqt:t-test}
		\sqrt{n} \frac{\mu}{\sigma} > 2.326
\end{equation}

\subsubsection{Z-test}
Equation~\ref{eqt:z-test} shows how we computed the two proportion z-test pooled
for $\mu_{pre}$ equal $\mu_{post}$. We first compute the $z_{score}$, then we
compare it with the Z-value that ensures confidence of 99\%~\cite{z-test, wiki:z-test}.
The $z_{score}$ is the observed difference ($a$) divided by the standard error of the
difference ($b$) pooled from the two samples ($p$).

\begin{equation}
		\label{eqt:z-test}
		\begin{split}
				& a = \mu_{pre} - \mu_{post}\\
				& p = \frac{\mu_{pre} + \mu_{post}}{2}\\
				& b = \sqrt{\frac{2 p (1-p) }{n}}\\
				& z_{score} = \frac{a}{b}\\
				& z_{score} > 2.576 
		\end{split}
\end{equation}

\subsubsection{Sum of Standard Deviations}
Equation~\ref{eqt:sum-stdv} shows the use of the sum of standard deviations as
comparison test. The difference between postquential and prequential is
significant when that difference is higher than the sum of
standard deviations.

\begin{equation}
		\label{eqt:sum-stdv}
		\mu_{post} - \mu_{pre} > \sigma_{post} + \sigma_{pre}
\end{equation}

\subsection{Pre- and Postquential Statistics Computation}
\label{sec:mean-variance}
In Algorithm~\ref{alg:training-forest} we mention that the accuracy and its
variance are evaluated both prequentially and postquentially. However, only the
most recent data points are relevant for these statistics. We compared two ways
of computing the mean ($\mu$) and variance ($\sigma^2$): sliding and fading.

The sliding version uses a sliding window to store the statistics. Let
$P_i \in \{0, 1\}$ be the correctness of the prediction for data point $i$. The
values of $P_i$ are stored in a binary sliding window $W$ of size $W_{size}$.
$W$ is updated with the most recent $P_i$ and the mean and variance of the accuracy are computed
as follows:
\begin{equation}
		\begin{split}
				& \mu = \frac{\sum_{P_i \in W}P_i}{W_{size}}\\
				& \sigma^2 = \mu(1-\mu)\\
		\end{split}
\end{equation}
This expression of $\sigma^2$ comes from the fact that $P_i$ is a binary
variable.

The sliding version increases memory consumption because it needs to keep $W$
in memory. The fading version addresses this downside of the sliding version.
To reduce memory usage, the fading version relies on a fading
factor~\cite{issues_learning_from_stream}. The sum maintained to compute the
accuracy and the variance are faded. Which mean these sums are multiplied by a
fading factor $f\in[0, 1]$ before being updated. It gives a weight to all
elements with older elements having a smaller weight. If $f=1$ then the sum is
computed for the entire stream. If $f=0$ then only the last point is taken into
account.

To compute the fading statistics, we need the count of data points $n$, the
faded count of data points $N = \sum_{i=1}^n f^{n-i}$, and the faded accuracy of
the prediction $A_n = \sum_{i=1}^n f^{n-i}P_i$. This is
sufficient to compute the mean accuracy and its variance:
\begin{equation}
		\begin{split}
				& \mu = \frac{A_n}{N}\\
				& \sigma^2 = \mu(1-\mu)\\
		\end{split}
\end{equation}
In this experiment, we use a fading factor of $0.995$.

\subsection{Tree Addition Method}
\label{sec:adding-tree}
In the Data Stream \mondrianforest, adding a tree simply implants a root in the
node pool. The main issue in adding trees revolves around the number of nodes
available for that tree to grow. Indeed, if the number of nodes available is too
small, the tree won't grow much and it will underfit the data.

Therefore, to make space for new trees, we need to trim the leaves of existing
trees. The trimming phase ensures that every tree has a similar size while
accommodating enough memory space for the new tree to grow. We test three
approaches for trimming trees: Add random, Add depth, and Add count. The Add
random approach randomly selects the leaves to remove. The Add depth approach
removes the deepest leaves first. The Add count approach focuses on the leaves
that contain the least amount of data points.

\subsection{Datasets}
\label{sec:datasets}
We used six datasets to evaluate our proposed methods: three synthetic datasets to mimic real-world
situations and to make comparisons with and without concept drifts, and two real
 Human Activity Recognition datasets.

\subsubsection{\banosdataset}
The \banosdataset~dataset~\cite{Banos_2014}\footnote{available
\href{https://archive.ics.uci.edu/ml/datasets/REALDISP+Activity+Recognition+Dataset\#:\~:text=The\%20REALDISP\%20(REAListic\%20sensor\%20DISPlacement,\%2Dplacement\%20and\%20induced\%2Ddisplacement}{here}}
is a human activity dataset with 17 participants and 9~sensors per participant.
Each sensor samples a 3D acceleration, gyroscope, and magnetic field, as well as
the orientation in a quaternion format, producing a total of 13 values.  Sensors
are sampled at 50~Hz, and each sample is associated with one of 33 activities.
In addition to the 33 activities, an extra activity labeled 0 indicates no
specific activity.

We pre-processed the \banosdataset dataset as in \cite{Banos_2014}, using non-overlapping windows of one
second (50~samples), and using only the 6 axes (acceleration and gyroscope) of
the right forearm sensor. We computed the average and standard deviation over
the window as features for each axis. We assigned the most frequent label to the
window. The resulting data points were shuffled uniformly.

In addition, we constructed another dataset from \banosdataset, in which we
simulated a concept drift by shifting the activity labels in the second half of
the data stream. 

\subsubsection{\recofitdataset}
The \recofitdataset dataset~\cite{recofit, recofit_data} is a human activity dataset
containing 94 participants.
Similar to the \banosdataset dataset, the activity labeled 0 indicates no
specific activity. Since many of these activities were similar, we merged  some
of them together based on the table in~\cite{cross_subject_validation}. 

We pre-processed the dataset similarly to the \banosdataset dataset, using
non-overlapping windows of one second, and only using 6 axes (acceleration and
gyroscope) from one sensor. From these 6 axes, we used the average and the
standard deviation over the window as features. We assigned the most frequent
label to the window.

\subsubsection{MOA Datasets}
We generated two synthetic datasets using Massive Online Analysis~\cite{moa}
(MOA) is a Java framework to compare data stream classifiers. In addition to
classification algorithms, MOA provides many tools to read and generate
datasets. We~generate two synthetic datasets (MOA commands available
\href{https://github.com/big-data-lab-team/benchmark-har-data-stream/blob/956e81f446a531111c1680a1abb96d6117a19a87/Makefile#L200}{here})
using the RandomRBF algorithm, a stable dataset, and a dataset with a drift.
Both datasets have 12 features and 33 labels, similar to the \banosdataset
dataset. We generated 20,000 data points for each of these synthetic datasets.

\subsubsection{Covtype}
The \covtypedataset dataset\footnote{available
\href{https://archive.ics.uci.edu/ml/datasets/covertype}{here}} is a tree dataset. Each data
point is a tree described by 54 features including ten quantitative variables
and 44 binary variables. The 581,012 data points are labeled with one of the seven forest cover
types and these labels are highly imbalanced. In particular, two labels represent
85\% of the dataset.

\subsection{Evaluation Metric}
\label{sec:evaluation-metric}
We evaluated our methods using a prequential fading macro F1-score. We focused
on the F1 score because most datasets are imbalanced. We used the prequential
version of the F1 score to evaluate classification on data
stream~\cite{issues_learning_from_stream}. We used a fading factor to minimize
the impact of old data points, especially data points at the beginning or data
points saw before a drift occur. To obtain this fading F1 score, we multiplied
the confusion matrix with the fading factor before incrementing the cell in the
confusion matrix.


\section{Results}
In this section, we first highlight the existence of an optimal number of trees
dependent on the data and the memory. Then we evaluate the performance of the
tree-adding methods independently from the dynamic update process.
Finally, we assess the complete dynamic update method and all its parameter.

\subsection{Optimal Forest Size}
\label{sec:result-optimal-tree-count}
\begin{figure}[!t]
\centering
	\includegraphics[scale=0.24]{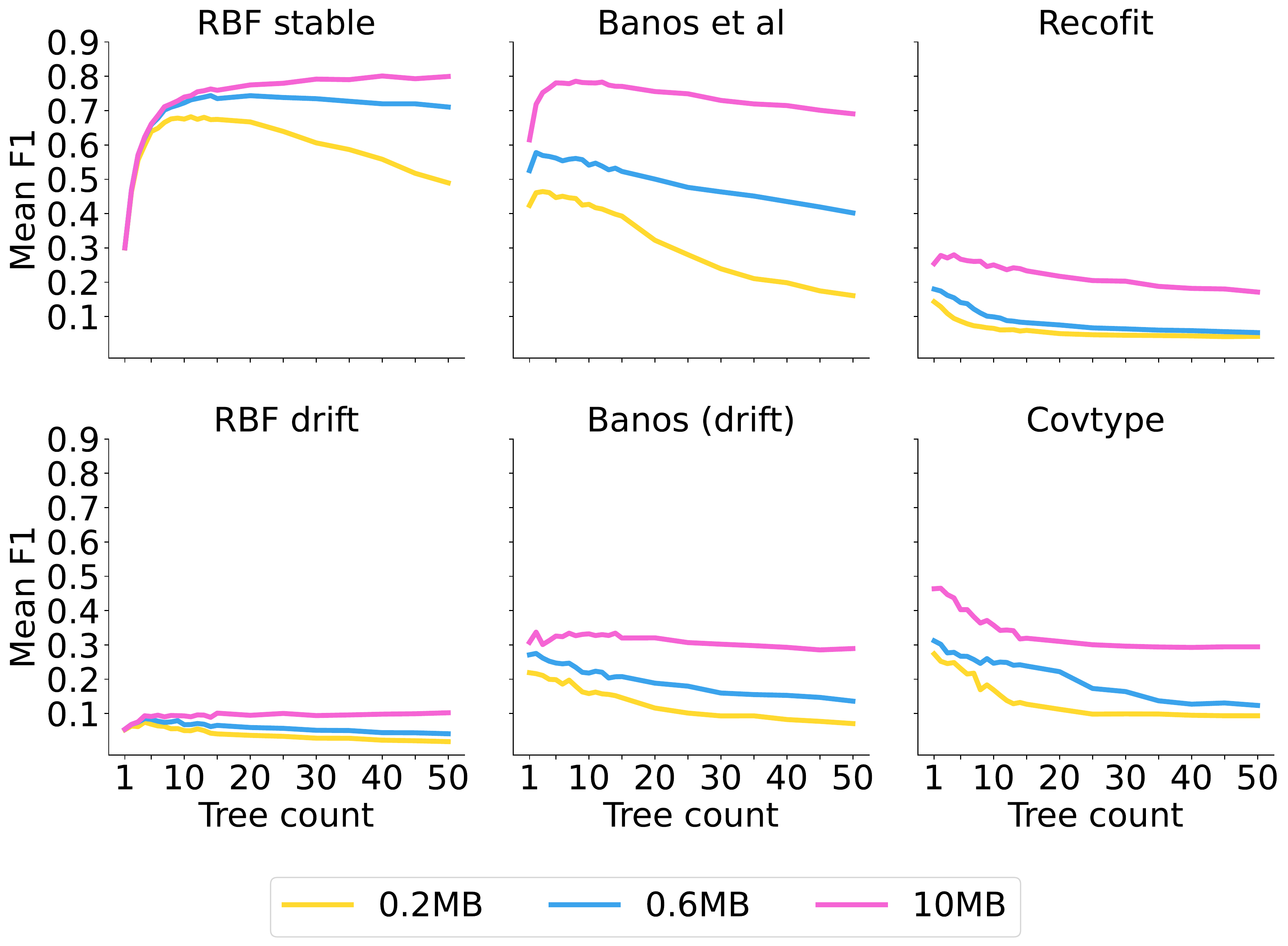}
	\caption{The impact of the ensemble size on the F1 score depending on the datasets and the memory
		limit.}
	\label{fig:f1_impact_tree_cout}
\end{figure}

Figure~\ref{fig:f1_impact_tree_cout} shows the relation between the number of
trees in the \mondrianforest and the F1 score. We notice that in most
configurations, there is an optimal number of trees located between 1 and 15,
except for 10MB and the datasets RBF stable and \banosdataset, in which case
the F1 score keeps increasing without reaching a maximum.

A particular situation occurs on the \covtypedataset and the \recofitdataset
datasets with 0.2MB and 0.6MB: the optimal ensemble size is one, which suggests
that the memory limit is not high enough for a tree to overfit the
datasets. Therefore, adding trees will always underfit.

We observe significant performance differences between the best-performing and
least-performing number of trees, in particular for low memory amounts.
Therefore, optimizing the number of trees is necessary to achieve the best
performances.

\subsection{Tree Addition}
\begin{figure}[!t]
\centering
	\includegraphics[scale=0.24]{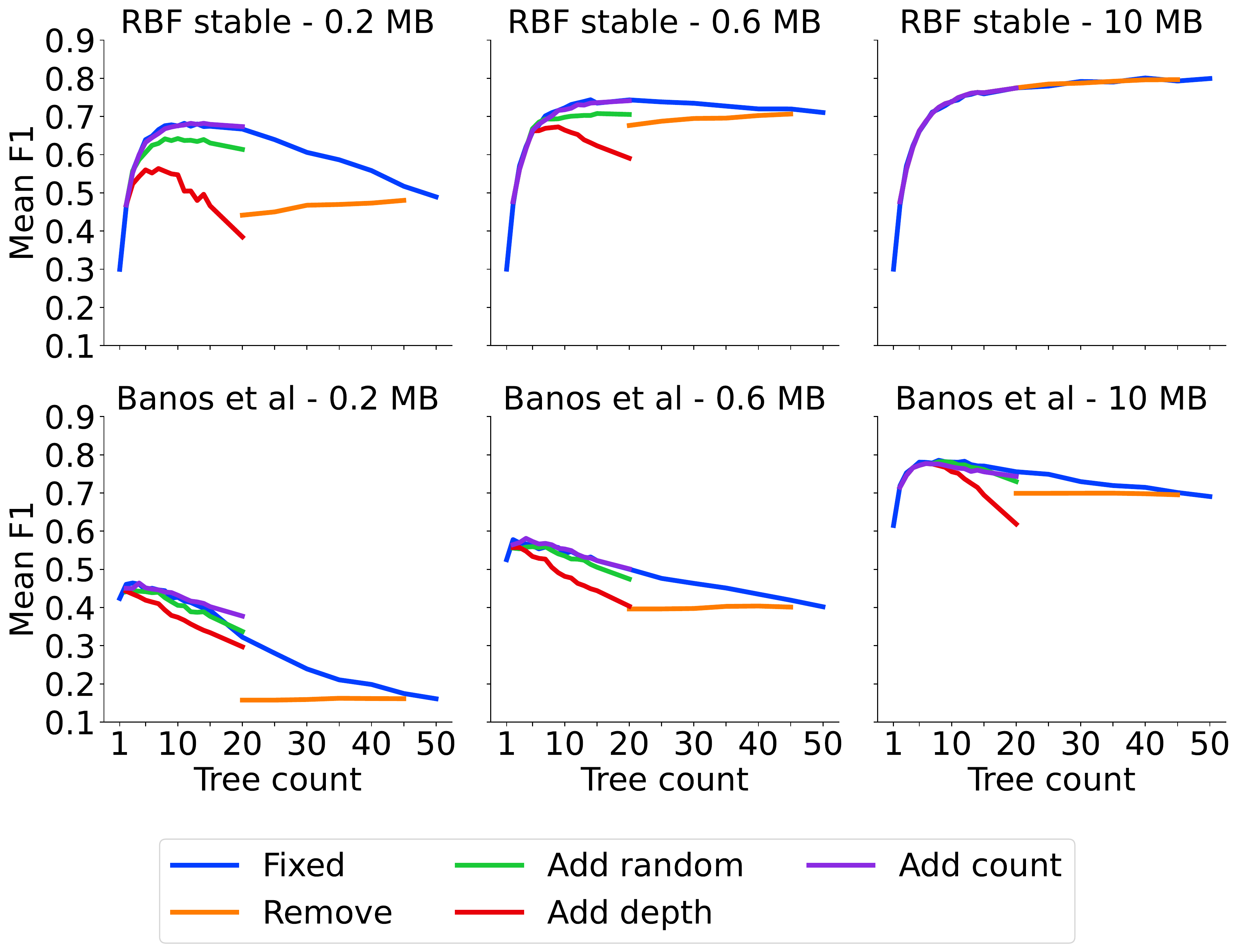}
	\caption{The effectiveness of the adding methods and removing method compare to Fixed.}
	\label{fig:f1_add_remove}
\end{figure}

Figure~\ref{fig:f1_add_remove} compares the tree addition methods presented in
Section~\ref{sec:adding-tree} with Fixed, the \mondrianforest with a fixed
number of trees described in Section~\ref{sec:result-optimal-tree-count}. This
comparison is done independently of the dynamic procedure that will be
evaluated later. In this Figure, Add random, Add count, and Add depth, start
their forest with 1 tree, then add a tree periodically until the forest reaches
the size given on the x-axis. Ideally, the update method should be
indistinguishable from Fixed that uses the number of trees indicated on the
x-axis.

From Figure~\ref{fig:f1_add_remove}, we note that Add Count is consistently
closer to Fixed than Add random is. Add depth tend to be the worst variant to
add a new tree as it diverges faster than the other two.
Therefore, adding trees with Add count is a functionnal strategy to dynamicly
adjust the number of tree.

\subsection{Tree Removal}
Since detecting underfitting is challenging, tree removal is not
included in Algorithm~\ref{alg:training-forest}. However, we implemented a tree
removal method that we evaluated similarly to the add method, in
Figure~\ref{fig:f1_add_remove}. The Remove method in that Figure starts a forest
with 50 trees and periodically removes a tree to reach the value indicated on
the x-axis.

We note that removing trees always underperforms except for the highest amounts
of memory. Indeed, removing a tree to decrease underfitting by allowing the
remaining trees to grow more nodes raises an issue related to outlier data
points. The forest deletes a tree when the memory limit has been reached.
Therefore, the tree growth has been paused and only the node statistics (box and
counters) are updated. Once a tree is deleted, the other trees resume their
growth until the memory is full again. However, during the pause phase, data
points have still been received, but since none of the points outside a box
could branch off, they forced the boxes to expand so they fit all data points.
Thus, the node boxes tend to be bigger.

When growth is resumed, data points that could introduce new nodes are more
likely to be outliers since only data points outside a box can branch off.
Therefore, most of the new nodes, if not all, will be created for outliers, and
these nodes are unlikely to be used and to improve the classification.

\subsection{Comparison to Fixed Ensemble Size}
We tested all possible combinations of:
\begin{itemize}
		\item Tree addition methods (Section~\ref{sec:adding-tree})
		\item Pre- and Postquential statistics (Section~\ref{sec:mean-variance})
		\item Comparison Tests (Section~\ref{sec:comparison-test})
\end{itemize}

The tree addition includes Add random, Add depth, and Add count. The
pre- and postquential statistics include fading and sliding. Finally, the comparison test
contains the sum of standard deviation (sum-std), the sum of variance (sum-var),
the t-test (t-test), and the z-test (z-test).

\begin{figure}[!t]
\centering
	\includegraphics[height=0.8\textheight]{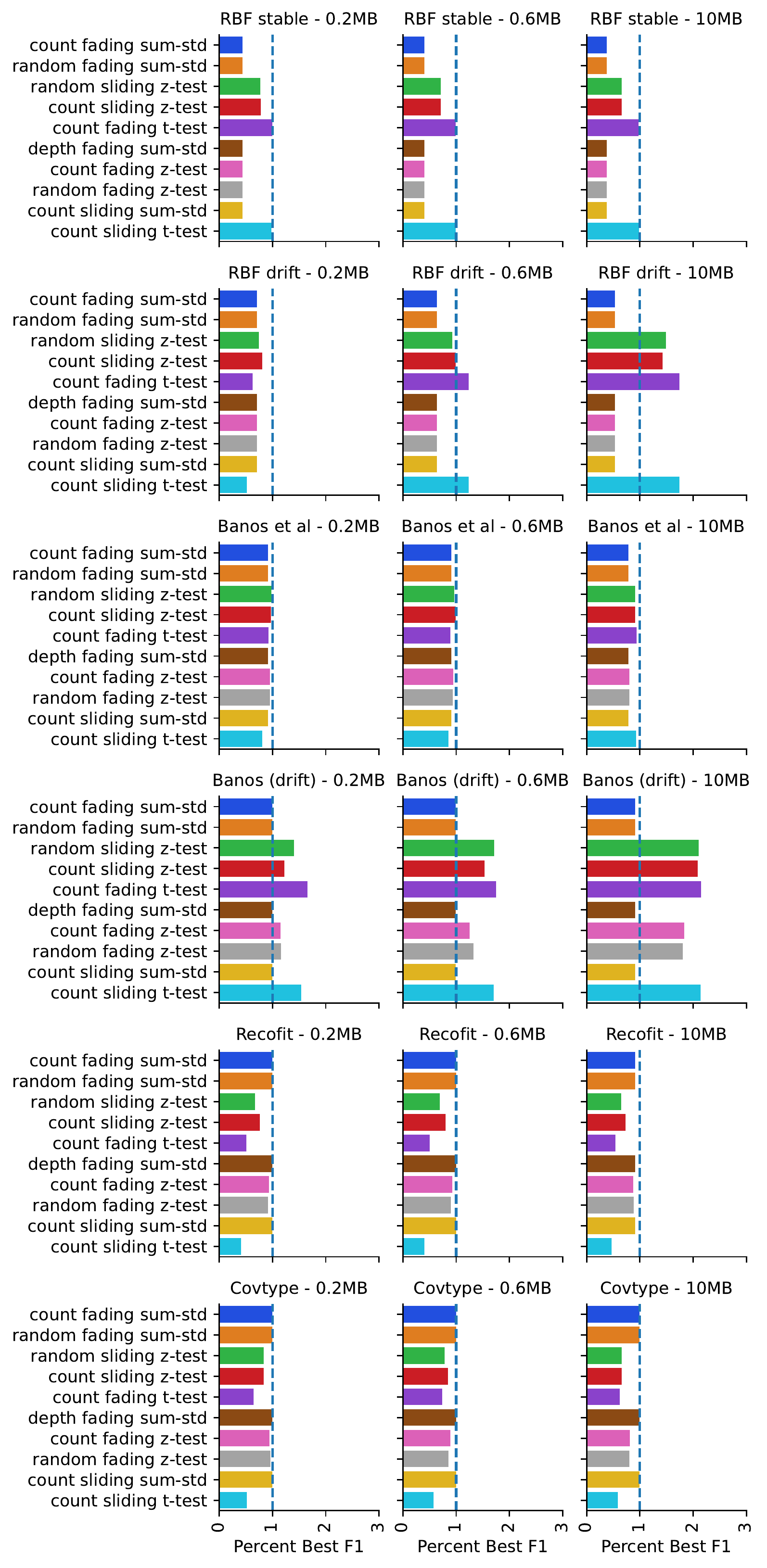}
	\caption{Comparison between component combinations and Fixed with the optimal
		tree count. Fixed is represented by a vertical dashed line. The value for
		the component combinations is percentage of the Fixed F1 score.}
	\label{fig:f1_xp3}
\end{figure}

Figure~\ref{fig:f1_xp3} shows how the top combinations compare to Fixed for
each dataset and memory limit. The score is relative to the F1 score of Fixed
with the optimal number of trees.

We observe that for most datasets, at least one dynamic forest reaches the
performance of the Fixed method. The only exception is the RBF drift dataset
with 0.2MB where all the dynamic approaches are substantially under the
performance obtained by the Fixed method.

For the drift datasets, some dynamic forests surpass the performance of the
Fixed method. This is due to the introduction of new trees that are not
influenced by older concepts, and thus are more accurate to the new data
points.

Nevertheless, no single dynamic method consistently reaches the performance of
Fixed. Indeed, the count fading t-test method (purple on
Figure~\ref{fig:f1_xp3}) reaches or surpasses the performance of the Fixed
method for RBF stable and drift (except 0.2MB), \banosdataset and \banosdataset
drift, however, it underperforms on the \recofitdataset and \covtypedataset
datasets where the count fading sum-std (blue on Figure~\ref{fig:f1_xp3}) and the random fading sum-std
(orange on Figure~\ref{fig:f1_xp3}) perform significantly better. Conversely, the depth fading sum-std
method (brown on Figure~\ref{fig:f1_xp3}) reaches the Fixed performance for \covtypedataset,
\recofitdataset, and \banosdataset datasets, but significantly underperforms on
RBF stable and RBF drift.

\begin{table}
		\centering
		\begin{tabular}{|c|c|c||c|}
				\hline
				Tree Addition & Pre \& Postquential & Comp. Test & Avg. Rank \\ 
				\hline
				count  & fading  & sum-std & 6.50 \\
				random & fading  & sum-std & 7.50 \\
				random & sliding & z-test  & 7.61 \\
				count  & sliding & z-test  & 7.67 \\
				count  & fading  & t-test  & 8.39 \\
				depth  & fading  & sum-std & 8.50 \\
				count  & fading  & z-test  & 8.67 \\
				random & fading  & z-test  & 9.28 \\
				count  & sliding & sum-std & 9.50 \\
				count  & sliding & t-test  & 10.44\\
				\hline
		\end{tabular}
		\caption{The best-ranked component combinations out of the 24 combinations of
		Algorithm~\ref{alg:training-forest} across datasets and memory limits. The
		top lines rank better on average than the bottom lines.}
		\label{tab:global-f1}
\end{table}

We computed the average rank of the dynamic forests and reported the top 10
methods in Table~\ref{tab:global-f1}. The best average rank of 6.50 out of 24
indicates the lack of a clear winner among all combinations.

Moreover, we observed that varying the fading factor made some methods
consistently approach the Fixed performance. In particular, when reducing the
fading factor, the methods using the z-test and the t-test approach Fixed on
all datasets. This is explained by the fact that the experimental conditions
are closer to the assumptions made by the t-test and the z-test.
Nevertheless, these fading factor explorations were not done with a proper
cross-validation, and cannot be repported in this paper because it would require
an analysis on independent datasets.


\section{Related Work}
\label{sec:related-work}
The work in~\cite{adjust_ensemble_size} proposes a method to adjust ensemble size
based on accuracy contribution of the sub-classifier. A sub-classifier that
significantly decreases the accuracy of the ensemble should be removed, whereas
the new sub-classifier built on the last chunk of data should be added if it
significantly increases the ensemble accuracy. However, their hypotheses are not
valid in our study, since growing a new tree influences the performance of
existing trees. Indeed, the new tree requires nodes to grow and these nodes will
be taken from existing trees.

The work in~\cite{hyper-parameter_tuning} explores an algorithm to adjust
hyper-parameters on data streams with concept drift. When a concept drift is
detected, the algorithm makes a list of hyper-parameters configurations to
evaluate on the most recent data points. The best configuration provides the
new hyper-parameters until the next concept drift. This method is not suited
for our hypothesis since it assumes we have enough memory to store at least two
models.


\section{Conclusion}
In this paper, we showed experimentally that a memory-bounded \mondrianforest
has an optimal ensemble size that depends on the dataset and the memory limit.
To find this optimal, we proposed a dynamic ensemble-sized \mondrianforest that
estimates overfitting to drive the ensemble toward the optimal number of trees.
The overfitting measure relies on the postquential accuracy, an innovative
concept to estimate the training accuracy of a data stream classifier.

We introduced the use of fading factor to keep track of the mean and variance,
needed for the comparison test.
We tested our algorithm on six datasets with different combinations of
comparison tests, different methods to add trees, and different ways of
computing the mean and variance.

From this experiment, we observed that some of the proposed methods were able
to reach the performance of a \mondrianforest with an optimal number of
trees. However, none of the methods consistently achieve that optimal.
In future work, we suggest investigating the role of the fading factor.

In addition, further investigations are required to design a functional tree
removal method that resists two issues: detecting underfitting in the forest
and continuing the growth of paused \mondriantrees.

\section*{Acknowledgement}
This work was funded by a Strategic Project Grant of the Natural Sciences
and Engineering Research Council of Canada. The computing platform was
obtained with funding from the Canada Foundation for Innovation.


\bibliographystyle{IEEEtran}
\bibliography{IEEEabrv, paper}
\end{document}